\documentclass[11pt]{article}

\usepackage[final]{acl}

\usepackage{tabularx}

\usepackage{times}
\usepackage{latexsym}

\usepackage[T1]{fontenc}

\usepackage[utf8]{inputenc}

\usepackage{microtype}

\usepackage{inconsolata}
\usepackage{microtype}

\usepackage{subcaption}
\usepackage{graphicx}
\usepackage{booktabs}
\usepackage{multirow}
\usepackage{amsmath}
\usepackage{algpseudocode}
\usepackage{amsfonts}
\usepackage{amssymb,graphicx}
\usepackage{makecell}
\usepackage{algorithm}
\usepackage{tabularx}
\usepackage{supertabular}

\usepackage[figuresleft]{rotating}
\usepackage{pifont}
\newcommand{\cmark}{\ding{51}}%
\newcommand{\xmark}{\ding{55}}%
\usepackage{ltablex}
\usepackage{csquotes}
\usepackage{tikz}
\usepackage{forest}
\usetikzlibrary{trees,positioning,shapes,shadows,arrows.meta}
\title{Table Question Answering in the Era of Large Language Models: \\A Comprehensive Survey of Tasks, Methods, and Evaluation}

\author{Wei Zhou$^{1,3}$ \hspace{3mm}
Bolei Ma$^2$ \hspace{3mm}
  Annemarie Friedrich$^3$\hspace{3mm}
    Mohsen Mesgar$^1$ 
    \\
  $^1$Bosch Center for Artificial Intelligence, Renningen, Germany \\ 
      $^2$LMU Munich \& Munich Center for Machine Learning, Germany \\  $^3$University of Augsburg, Germany \hspace{5mm} \\
\texttt{\{wei.zhou3|mohsen.mesgar\}@de.bosch.com}\\ 
  \texttt{bolei.ma@lmu.de} \hspace{3mm}
\texttt{annemarie.friedrich@uni-a.de}}

\begin{document}
\maketitle
\begin{abstract}
Table Question Answering (TQA) aims to answer natural language questions about tabular data, often accompanied by additional contexts such as text passages.
The task spans diverse settings, varying in table representation, question/answer complexity, modality involved, and domain.
While recent advances in large language models (LLMs) have led to substantial progress in TQA, the field still lacks a systematic organization and understanding of task formulations, core challenges, and methodological trends, particularly in light of emerging research directions such as reinforcement learning.
This survey addresses this gap by providing a comprehensive and structured overview of TQA research with a focus on LLM-based methods. 
We provide a comprehensive categorization of existing benchmarks and task setups. 
We group current modeling strategies according to the challenges they target, and analyze their strengths and limitations. 
Furthermore, we highlight underexplored but timely topics that have not been systematically covered in prior research.
By unifying disparate research threads and identifying open problems, our survey offers a consolidated foundation for the TQA community, enabling a deeper understanding of the state of the art and guiding future developments in this rapidly evolving area.

\end{abstract}





\section{Introduction}

Tables are a ubiquitous data format in daily life \cite{10.14778/1453856.1453916}. Automatically processing and understanding tabular data with (multi-modal) large language models ((M)LLMs) has recently attracted considerable attention from both industry \cite{katsis-etal-2022-ait, Su2024TableGPT2AL} and academia \cite{pasupat-liang-2015-compositional, Wolff2025HowWD}, emerging as a prominent research direction.

\begin{figure}[t]
    \centering
    \includegraphics[width=1\linewidth]{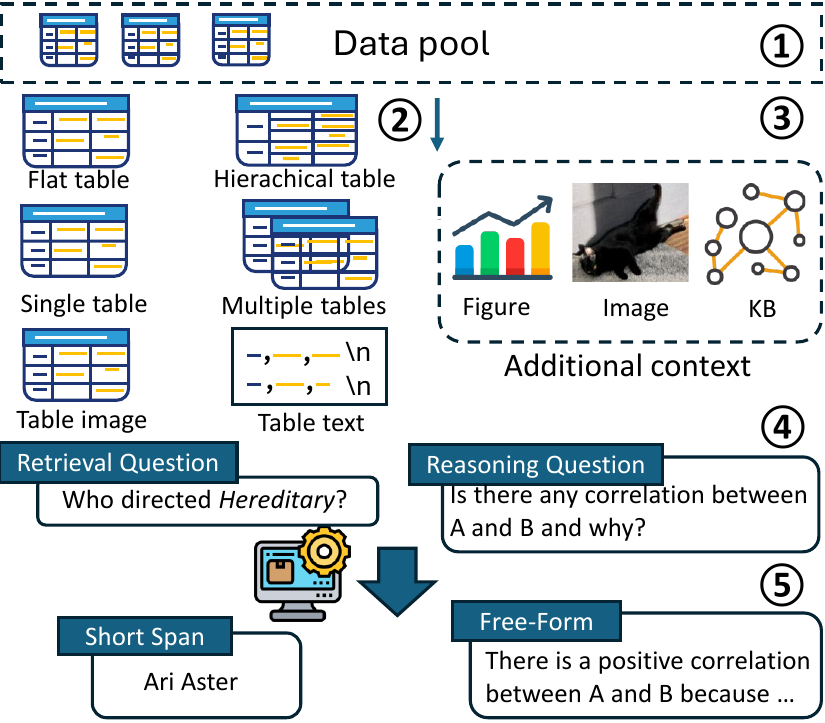}
    \caption{Five table question answering task setups.\ding{172} \textbf{Domain:} Either the inputs need to be retrieved from a data pool (open-domain) or directly given (closed-domain). \ding{173} \textbf{Table}: Input tables can vary in structure, numbers, and representations. \ding{174} \textbf{Additional Context}: Charts, images, and knowledge graphs can also be involved as inputs. \ding{175} \textbf{Question Complexity}: A question can involve retrieving certain cells from a table or require reasoning and analysis to be solved. \ding{176} \textbf{Answer Format}: Answers can be in short text spans, consisting only of numbers and entities, or in free-form natural language, with no limitation on types and length.}
    \label{fig:tqa_tasks}
\end{figure}

Among the various tasks involving tables, including table generation \cite{NEURIPS2023_90debc7c} and table-to-text \cite{parikh-etal-2020-totto}, table question answering (TQA) stands out as one of the most widely studied \cite{wu2025tabulardataunderstandingllms}. The goal of TQA is to answer questions based on tabular data, optionally augmented with additional context such as text passages or images. 
TQA can be instantiated in diverse settings. As illustrated in Figure~\ref{fig:tqa_tasks}, the input table may be provided directly or retrieved from a large corpus; tables vary in format, size, and structural complexity; and questions may target specific cells or require multi-step reasoning. These variations reflect real-world applications and necessitate different modeling strategies, driving rapid growth in the field. This survey consolidates resources and modeling approaches, and distills insights into promising directions for future research.

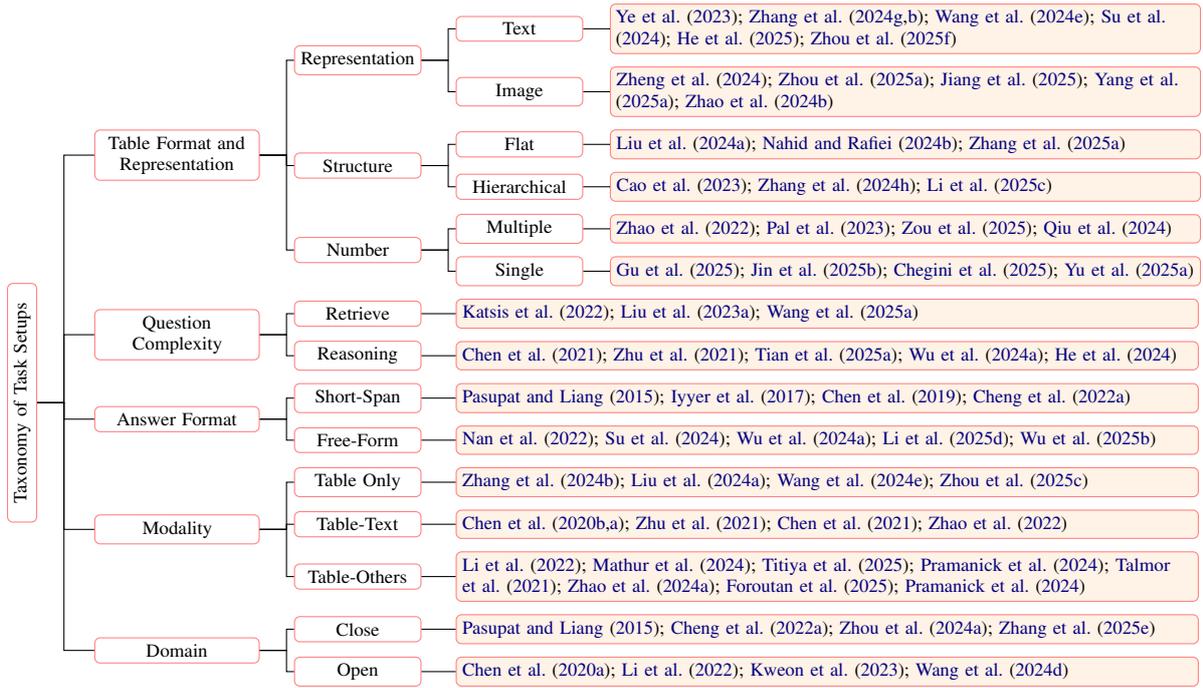
\begin{figure*}[htbp]
\scriptsize
%
\tikzset{  
    basic/.style  = {draw, text width=1.45cm, align=center, rectangle},  
    root/.style   = {basic, rounded corners=2pt, thin, align=center, fill=white, draw=red!50},  
    xnode/.style  = {basic, thin, rounded corners=2pt, align=center, fill=white, draw=red!50, text width=2cm, anchor=center}, 
    pnode/.style  = {basic, thin, rounded corners=2pt, align=center, fill=white, draw=red!50, text width=1.5cm, anchor=center},  
    anode/.style  = {basic, thin, rounded corners=2pt, align=left, fill=orange!10, draw=red!50, text width=7.60cm, anchor=center},  
    bnode/.style  = {basic, thin, rounded corners=2pt, align=left, fill=orange!10, draw=red!50, text width=9.6cm, anchor=center}, 
    edge from parent/.style={draw=black, edge from parent fork right}  
}  
\begin{forest} 
for tree={  
    grow=east,  
    growth parent anchor=east,  
    parent anchor=east,  
    child anchor=west,  
    edge path={  
        \noexpand\path[\forestoption{edge}]  
        (!u.parent anchor) -- +(10pt,0) |- (.child anchor) \forestoption{edge label};  
    },  
}  
[Taxonomy of Task Setups, root, l sep=7.5mm, rotate=90, child anchor=north, parent anchor=south, anchor=center, text width=3cm, 
    [Domain, xnode, l sep=4.5mm, 
        [Open, pnode, l sep=4.5mm, 
            [\citet{Chen2020OpenQA, li-etal-2022-mmcoqa, kweon-etal-2023-open,Wang2024RETQAAL}, bnode]
        ]
        [Closed, pnode, l sep=4.5mm,
           [\citet{pasupat-liang-2015-compositional,cheng-etal-2022-hitab,  zhou-etal-2024-freb,zhang-etal-2025-scitat}, bnode]
        ]
    ]  
     [Modality, xnode, l sep=4.5mm,  
        [Table-Others, pnode, l sep=4.5mm,
             [\citet{li-etal-2022-mmcoqa,mathur-etal-2024-knowledge,Titiya2025MMTBENCHAU,Pramanick2024SPIQAAD, Talmor2021MultiModalQACQ,10.1145/3664647.3681053,Foroutan2025WikiMixQAAM,Pramanick2024SPIQAAD}, bnode]
        ]
        [Table-Text, pnode, l sep=4.5mm, 
             [\citet{chen-etal-2020-hybridqa, Chen2020OpenQA, zhu-etal-2021-tat, chen-etal-2021-finqa, zhao-etal-2022-multihiertt}, bnode]
        ]
        [Table Only, pnode, l sep=4.5mm, 
             [\citet{zhang-etal-2024-tablellama,liu-etal-2024-rethinking,Wang2024ChainofTableET,Zhou2025PPTAP}, bnode]
        ]
     ]  
     [Answer Format, xnode, l sep=4.5mm,  
        [Free-Form, pnode, l sep=4.5mm, 
             [\citet{nan-etal-2022-fetaqa, Su2024TableGPT2AL,Wu2024TableBenchAC,li-etal-2025-mimotable,wu-etal-2025-realhitbench}, bnode]
        ]   
        [Short-Span, pnode, l sep=4.5mm,
             [\citet{pasupat-liang-2015-compositional,iyyer-etal-2017-search,Chen2019TabFactAL, cheng-etal-2022-hitab}, bnode]
        ]
     ]  
     [Question Complexity, xnode, l sep=4.5mm,  
        [Reasoning, pnode, l sep=4.5mm, 
             [\citet{chen-etal-2021-finqa,zhu-etal-2021-tat,tian2025realworldtableagentscapabilities,Wu2024TableBenchAC,He2023Text2AnalysisAB}, bnode]
        ]   
        [Retrieve, pnode, l sep=4.5mm,
             [\citet{katsis-etal-2022-ait,liu-etal-2023-tab,Wang2025NeedleInATableEL}, bnode]
        ]
     ]  
     [Table, xnode, l sep=4.5mm,  
        [Number, pnode, l sep=4.5mm, 
            [Single, pnode, l sep=3.5mm,
                 [\citet{gu-etal-2025-toward,Jin2025Tabler1SA, chegini-etal-2025-repanda, yu-etal-2025-table}, anode]
            ]
            [Multiple, pnode, l sep=3.5mm, 
                 [\citet{zhao-etal-2022-multihiertt, pal-etal-2023-multitabqa,Zou2025GTRGF, Qiu2024TQABenchEL}, anode]
            ]
        ]
        [Structure, pnode, l sep=4.5mm, 
            [Hierarchical, pnode, l sep=3.5mm,
                 [\citet{cao-etal-2023-api,zhang-etal-2024-e5,li-etal-2025-graphotter}, anode]
            ]
            [Flat, pnode, l sep=3.5mm, 
                 [\citet{liu-etal-2024-rethinking,Nahid2024TabSQLifyER, zhang-etal-2025-alter}, anode]
            ]
        ]  
        [Representation, pnode, l sep=4.5mm,
            [Image, pnode, l sep=3.5mm,
                 [\citet{zheng-etal-2024-multimodal,11093154, Jiang2025MultimodalTR, Yang2025DoesTS, Zhao2024TabPediaTC}, anode]
            ]
            [Text, pnode, l sep=3.5mm, 
                 [\citet{10.1145/3539618.3591708, 10.14778/3659437.3659452, zhang-etal-2024-tablellama, Wang2024ChainofTableET, Su2024TableGPT2AL, he-etal-2025-tablelora, zhou-etal-2025-g}, anode]
            ]  
        ]
     ]  
]
\end{forest}
\caption{A taxonomy of TQA task setups. We list representative papers for each setup.}
\label{fig:taxonomy_task}
\end{figure*}

\paragraph{\textbf{Comparing with Existing Surveys.}} Most prior surveys relevant to TQA focus solely on textual tables \cite{ijcai2022p761,jin2022surveytablequestionanswering,Zhang2024ASO,fang2024largelanguagemodelsllmstabular,Ren2025DeepLW,XU2026114459}. Among those addressing both textual and image tables, \citet{wu2025tabulardataunderstandingllms} concentrate on table representations without discussing modeling approaches, while \citet{tian2025realworldtableagentscapabilities} emphasize agentic setups and overlook fine-tuning methods. Crucially, no existing survey provides an overview of TQA task setups or covers LLM-era advances such as reinforcement learning, interpretability, and novel evaluation paradigms (see detailed comparisons in Appendix~\ref{compare_survey}). 
\paragraph{\textbf{Scope.}} We include both TQA and table fact verification (TFV), in which models must determine if a given statement is valid based on tabular data. TFV can be reformulated as TQA \cite{lu-etal-2023-scitab}. We also incorporate Text-to-SQL datasets where applicable, as SQL queries can produce final answers. Our survey covers 277 papers published mostly after 2022, when LLMs were used. Details on the collection and paper statistics are in Appendix~\ref{survey_detail}.
\paragraph{\textbf{Structure.}} Section~\ref{sec:task_dimensions} outlines TQA task setups and benchmarks. Section~\ref{sec:Modeling_challenges} presents modeling approaches grouped by challenge. Section~\ref{sec:evaluation} reviews evaluation methodologies, and Section~\ref{sec:discussions} discusses emerging topics and future research directions.

\begin{figure*}[htbp]
\scriptsize
%
\tikzset{
     basic/.style  = {draw, text width=1.45cm, align=center, 
     rectangle, fill=green!10},
     root/.style   = {basic, rounded corners=2pt, thin, align=center, fill=white,draw=red!50, rotate=0},
     xnode/.style = {basic, thin, rounded corners=2pt, align=center, fill=white,draw=red!50,text width=2cm, anchor=center}, 
     pnode/.style = {basic, thin, rounded corners=2pt, align=center, fill=white,draw=red!50, text width=1.5cm, anchor=center},
     anode/.style = {basic, thin, rounded corners=2pt, align=left, fill=orange!10, draw=red!50, text width=11.5cm, anchor=center},
     bnode/.style = {basic, thin, rounded corners=2pt, align=left, fill=orange!10, draw=red!50, text width=9.3cm, anchor=center},
     edge from parent/.style={draw=black, edge from parent fork right}
}
\begin{forest} 
for tree={  
    grow=east,  
    growth parent anchor=east,  
    parent anchor=east,  
    child anchor=west,  
    edge path={  
        \noexpand\path[\forestoption{edge}]  
        (!u.parent anchor) -- +(10pt,0) |- (.child anchor) \forestoption{edge label};  
    },  
}  
[Taxonomy of Modeling Challenges, root, l sep=7.5mm, rotate=90, child anchor=north, parent anchor=south, anchor=center, text width=3cm, 
     [Knowledge Integration \S\ref{sec:knowledge-integration}, xnode, l sep=5.5mm, [\citet{Cheng2022BindingLM,Liu2024AugmentBY,Zhou2024EfficientMC, sui-etal-2024-tap4llm}, anode]  
     ]  
     [Data Heterogeneity \S\ref{sec:data-heterogeneity}, xnode, l sep=5.5mm,  
         [\citet{hu-etal-2024-ket, 10.1145/3664647.3681053, Liu2023MMHQAICLMI, li-etal-2022-mmcoqa,Chen2025PandoraAC, 10.1609/aaai.v39i24.34787, 10.1145/3616855.3635777}, anode]  
     ]  
     [Large Inputs \S\ref{sec:large-inputs}, xnode, l sep=5.5mm,  
        [Multi Tables, pnode, l sep=5.5mm,   [\citet{Kong2024OpenTabAL,liang-etal-2025-improving-table}, bnode]
          ] 
        [Large Table, pnode, l sep=5.5mm,
             [\citet{lin-etal-2023-inner, Lee2024PieceOT, Lee2024LearningTR, kojima-2024-sub, Patnaik2024CABINETCR}, bnode]
          ]
     ]  
     [Complex Query \S\ref{sec:complex_query}, xnode, l sep=5.5mm,  
          [Tuning-Free, pnode, l sep=5.5mm,
             [\citet{cheng-etal-2024-call, 10.14778/3659437.3659452, Lu2024TARTAO, Zhou2024EfficientMC, Yu2025TableRAGAR, Khoja2025WeaverIS, Cao2025TableMasterAR, 10.1145/3696410.3714962}, bnode]
          ]
        [Tuning-Base, pnode, l sep=5.5mm, 
             [\citet{zhang-etal-2024-tablellama, Xing2024TableLLMSpecialistLM, deng-mihalcea-2025-rethinking,Zhang2024TableLLMET, 10.1145/3654979,zheng-etal-2025-tabledreamer, chegini-etal-2025-repanda}, bnode]
          ]  
     ]  
     [Table Understanding \S\ref{sec:table_understanding}, xnode, l sep=5.5mm, 
         [Textual, pnode, l sep=5.5mm, 
             [\citet{chegini-etal-2025-repanda, zhou-etal-2025-g, Ye2024DataFrameQA,he-etal-2025-tablelora, mouravieff-etal-2025-structural, Su2024TableGPT2AL}, bnode]
          ]
         [Visual, pnode, l sep=5.5mm,
             [\citet{Nguyen2023TabIQATQ,HormazbalLagos2025ExpliCITQAEC,Zhao2024TabPediaTC, zheng-etal-2024-multimodal, 11093154, Jiang2025MultimodalTR, Yang2025DoesTS, Zhang2025TableMoENR}, bnode]
          ]
    ]   
]
\end{forest}
\caption{A taxonomy of methods categorized by challenges. We list representative papers for each challenge.}
\label{fig:taxonomy_challenge}
\end{figure*}
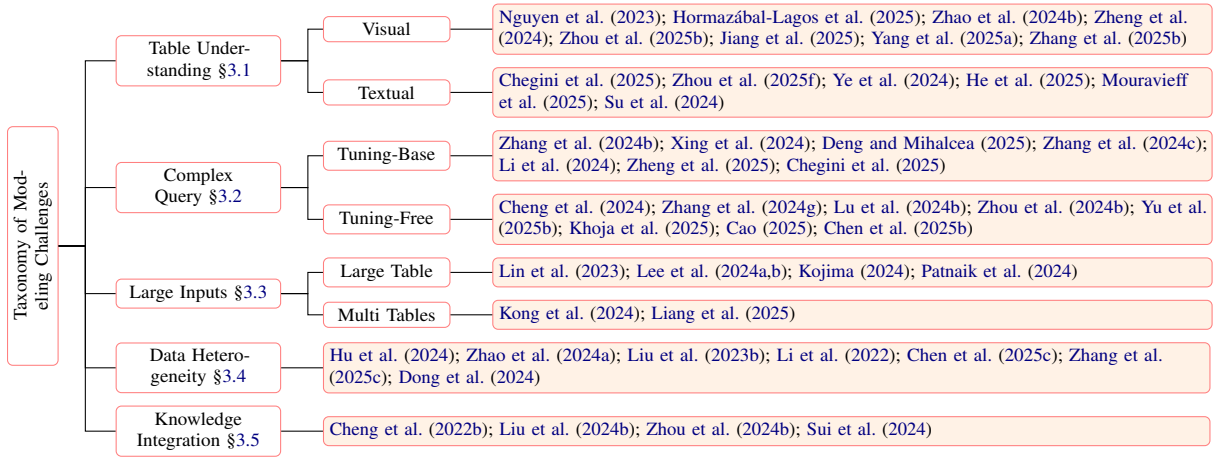

\section{Task Setups and Resources}
\label{sec:task_dimensions}
We dissect TQA from five perspectives. These are illustrated in Figure~\ref{fig:taxonomy_task}. Table \ref{tab:mapping_task_dataset_full} provides existing TQA datasets categorized by the characteristics of their task setups.

\paragraph{Table.} Tables may appear as text or images, giving rise to two distinct task setups: those operating over textual tables \cite{10.14778/3659437.3659452,zhang-etal-2024-tablellama,Wang2024ChainofTableET,Su2024TableGPT2AL,he-etal-2025-tablelora,zhou-etal-2025-g} and those over table images \cite{zheng-etal-2024-multimodal,11093154,Jiang2025MultimodalTR,Yang2025DoesTS,Zhao2024TabPediaTC}. Beyond format, table structure and quantity further shape task difficulty. Hierarchical tables \cite{cao-etal-2023-api,zhang-etal-2024-e5,li-etal-2025-graphotter} pose greater structural challenges than flat ones \cite{liu-etal-2024-rethinking,Nahid2024TabSQLifyER,zhang-etal-2025-alter}. Multi-table settings \cite{zhao-etal-2022-multihiertt,pal-etal-2023-multitabqa,Zou2025GTRGF,Qiu2024TQABenchEL} additionally require understanding inter-table relationships and handling longer inputs, compared to single-table setups \cite{gu-etal-2025-toward,Jin2025Tabler1SA,chegini-etal-2025-repanda,yu-etal-2025-table}.

\paragraph{Question Complexity.} TQA questions vary in the capabilities they require. Following \citet{zhou-etal-2024-freb}, they can be divided into \textit{retrieval} questions, solvable by locating relevant cells, and \textit{reasoning} questions, which demand additional inference. Accordingly, TQA benchmarks range from simple setups involving only retrieval \cite{katsis-etal-2022-ait,liu-etal-2023-tab,Wang2025NeedleInATableEL} to complex ones requiring numerical \cite{chen-etal-2021-finqa,zhu-etal-2021-tat,Lu2022DynamicPL,tian2025realworldtableagentscapabilities}, commonsense \cite{zhang-etal-2023-crt}, temporal \cite{gupta-etal-2023-temptabqa,shankarampeta-etal-2025-transienttables} reasoning, data analysis and plotting \cite{Wu2024TableBenchAC,He2023Text2AnalysisAB}. More recently, \citet{shen2025probabilisticquestionansweringtabular} introduce probabilistic TQA, requiring models to reason under uncertainty.

\paragraph{Answer Formats.} Most TQA setups feature short span answers composed of a few words or numbers \cite{pasupat-liang-2015-compositional,iyyer-etal-2017-search,Chen2019TabFactAL,cheng-etal-2022-hitab,wu2025mmqa}, which allow straightforward evaluation by exact match against reference answers. \citet{oses-grijalba-etal-2024-question} further categorize answers by type (e.g., Boolean, numerical), finding that Boolean questions tend to be the easiest. However, real-world queries often demand free-form answers spanning multiple sentences, as in data analysis tasks. This format is receiving growing attention for its alignment with practical use cases \cite{nan-etal-2022-fetaqa,Su2024TableGPT2AL,Wu2024TableBenchAC,li-etal-2025-mimotable,wu-etal-2025-realhitbench}.

\paragraph{Modality.} Beyond the standard table-question setup \cite{zhang-etal-2024-tablellama,liu-etal-2024-rethinking,Wang2024ChainofTableET,Zhou2025PPTAP}, many works introduce additional input modalities such as passages \cite{chen-etal-2020-hybridqa,Chen2020OpenQA,zhu-etal-2021-tat,chen-etal-2021-finqa,zhao-etal-2022-multihiertt}, images \cite{li-etal-2022-mmcoqa,mathur-etal-2024-knowledge,Titiya2025MMTBENCHAU,Pramanick2024SPIQAAD,Talmor2021MultiModalQACQ}, charts \cite{10.1145/3664647.3681053,Foroutan2025WikiMixQAAM,Pramanick2024SPIQAAD}, and knowledge graphs \cite{Christmann2023CompMixAB,hu-etal-2024-ket,huang-etal-2025-structfact}, better reflecting real-world settings where tables co-occur with heterogeneous data. Among these, the combination of tables and passages (table-text QA) is the most widely studied.

\paragraph{Domains.} TQA can be divided into open-domain \cite{Chen2020OpenQA,li-etal-2022-mmcoqa,kweon-etal-2023-open,strich-etal-2026-t2} and closed-domain \cite{pasupat-liang-2015-compositional,cheng-etal-2022-hitab,zhou-etal-2024-freb,zhang-etal-2025-scitat} settings, depending on whether the relevant inputs are provided. In open-domain TQA, a system must first retrieve relevant tables, either textual \citep{Chen2020OpenQA} or image-based \citep{xu-etal-2026-efficient}, from a large pool before reasoning over them, posing additional challenges compared to the closed-domain setting where target inputs are given directly.

         
         

\section{Modeling}
\label{sec:Modeling_challenges}
We categorize modeling methods based on the challenges they try to address. In addition, we also analyze their strengths and limitations. Figure~\ref{fig:taxonomy_challenge} provides an overview.

\subsection{Table Understanding}
\label{sec:table_understanding}
\paragraph{Visual Table Modeling.}
Visual table understanding requires comprehending both table content and structure. A common approach is to \textit{parse table images into text} using MLLMs \cite{Nguyen2023TabIQATQ,HormazbalLagos2025ExpliCITQAEC,si2025tabrag}. However, \citet{xia-etal-2024-vision} find that despite promising OCR performance on tabular data, MLLMs still struggle with spatial and formatting recognition, particularly for large or structurally complex tables \cite{zheng-etal-2024-multimodal,10.1007/978-3-032-09371-4_1}.

Another line of work focuses on \textit{pre-training and/or fine-tuning MLLMs} \cite{Zhao2024TabPediaTC,zheng-etal-2024-multimodal,11093154,Jiang2025MultimodalTR,Yang2025DoesTS,Zhang2025TableMoENR,yu2026thinkingtablesenhancingmultimodal}. Training datasets of table images have been constructed from existing tabular datasets \cite{Zhao2024TabPediaTC}, by converting textual tables into images \cite{zheng-etal-2024-multimodal,11093154,Jiang2025MultimodalTR}, or built from scratch \cite{Yang2025DoesTS,Zhang2025TableMoENR}. 

In terms of training strategies, \citet{zhu2026decouplingskeletonfleshefficient} decouple structure learning from content grounding during training. \citet{Kim2025TabFlashET} enhance visual table encoding through progressive question conditioning: injecting questions into vision transformers, along with a pruning mechanism to remove redundant background regions and a training recipe to mitigate the resulting information loss. Reinforcement learning has also been shown to improve performance in this setting \citep{kang-etal-2025-grpo}.
In terms of model architecture, dual vision encoders are employed to capture information at different levels of granularity \cite{Zhao2024TabPediaTC,11093154}, and \citet{Zhang2025TableMoENR} train a mixture-of-experts model to jointly capture layout and semantic information.

\paragraph{Textual Table Modeling.}
Prior work has explored three main directions for table structure understanding: table representations, architecture modifications, and specialized training tasks.
For \textit{representation}, tables have been modeled as relational databases or Pandas DataFrames, where operations are expressed in code \cite{chegini-etal-2025-repanda, zhou-etal-2025-g}, or as spreadsheets with formula-based operations \cite{Cao2025FortuneFR, Wang2025GeneralTQ}.
Compared to database and DataFrame formats, which require a pre-defined schema, spreadsheet representations offer greater flexibility.
Recent work improves the generalization of database representations by providing a processing module that converts a raw table into a SQL-ready table \cite{najpande2026quiettqueryindependenttabletransformation}.
Another approach models tables as graphs \citep{Hu2020HeterogeneousGT,NEURIPS2023_66178bea, Huang2025HyperGHL, li2025injecting, Jin2024HeGTaLH,wang2026linearizationattributedtablegraphs}, where nodes represent cells and edges encode positional relationships, enabling explicit structural learning.
For instance, \citet{Jin2024HeGTaLH} model tables as heterogeneous graphs with four node types: \texttt{TABLE} nodes, \texttt{ROW} nodes, \texttt{Header} nodes and \texttt{Data} nodes. 
Edges are defined based on node relationships, such as \texttt{Table–Header}.
Tables have also been expressed as natural language tuples \cite{zhao-etal-2023-large, yang2025triples} or trees \citep{wang2021tuta,cao2026orthogonalhierarchicaldecompositionstructureaware,10.1145/3769829,qu2026straptoragenticsemistructuredtable} 

Some methods \textit{encode structure within the model} \cite{he-etal-2025-tablelora, mouravieff-etal-2025-structural, Su2024TableGPT2AL}. For instance, \citet{he-etal-2025-tablelora} serialize tables with special tokens and applies 2D LoRA \cite{Hu2021LoRALA} to capture low-rank positional information, while \citet{mouravieff-etal-2025-structural} design a sparse attention mask for tabular data.

Finally, \textit{task-specific objectives} have been proposed to enhance structural reasoning, such as layout transformation inference \cite{Jin2025Tabler1SA}, where models detect changes between original and altered layouts, and cell position generation \cite{10936867}, where models predict cell locations.

\paragraph{Discussion.}
Visual table understanding is generally more challenging than textual table understanding \cite{deng-etal-2024-tables, zheng-etal-2024-multimodal}, possibly because it requires additional content interpretation. 
Notably, OCR-based pipelines do not outperform models directly fine-tuned on table images \cite{zheng-etal-2024-multimodal}.  For details, see Appendix \ref{method_compare}.
For textual table understanding, the need to define a fixed table schema may be a drawback for database tables. When using more fine-grained textual representations such as JSON or Markdown, there seems to be no optimal format across datasets and models \cite{Zhang2024FLEXTAFET}.


\subsection{Complex Query}
\label{sec:complex_query}

\paragraph{Tuning-Based.}
Methods in this group fine-tune (M)LLMs to improve reasoning over tabular data. Diverse training datasets covering a wide range of reasoning types are crucial for handling complex queries, and are either collected from existing benchmarks \cite{zhang-etal-2024-tablellama,Xing2024TableLLMSpecialistLM,deng-mihalcea-2025-rethinking} or synthesized \cite{Zhang2024TableLLMET,10.1145/3654979,zheng-etal-2025-tabledreamer,chegini-etal-2025-repanda}. For instance, \citet{zheng-etal-2025-tabledreamer} select training samples based on identified model weaknesses and progressively fine-tune the model, while \citet{chegini-etal-2025-repanda} distill Python programs generated by large closed-source models to train a smaller open-weight LLM. Combining fine-tuned models with tools at inference time has also been explored \cite{Wu2024ProTrixBM,mouravieff-etal-2024-learning,Vinayagame2024MATATAWS}, and reinforcement learning can further boost performance \cite{Nahid2024TabSQLifyER,Zhou2025PPTAP,Stoisser2025SparksOT,chi2025jtdaenhancingdataanalysis,zhang2026stareffectivestabletable,10.1145/3773966.3777932,li2026replacingmultistepassemblydata}.

Rather than fine-tuning the policy model, some work instead fine-tune a reward model that ranks candidate solutions \citep{zou2025tattootoolgroundedthinkingprm,tang-etal-2026-exploring}. Specifically, a process reward model is trained to score each intermediate reasoning step, with step scores aggregated for solution ranking and optionally used to guide next-step search. An alternative use of training data is to extract reusable experience from it \citep{gao2025utilizingtrainingdataimprove}, which can then be retrieved and appended to test instances at inference time.

\paragraph{Tuning-Free.}
In tuning-free approaches, agentic workflows that integrate tools are widely adopted to enable complex query handling and to reduce hallucinations \cite{cheng-etal-2024-call, 10.14778/3659437.3659452, Lu2024TARTAO, Zhou2024EfficientMC, Yu2025TableRAGAR, Khoja2025WeaverIS, Cao2025TableMasterAR, 10.1145/3696410.3714962, cao2026tablemaster}. 
These methods typically employ multi-step reasoning to decompose complex questions into simpler sub-problems \cite{Mao2024PoTableTS, Zhou2024EfficientMC, Ji2024TreeofTableUT, Zhang2024ALTERAF, Deng2024EnhancingTU, zhao-etal-2024-tapera, Nguyen2025PlanningFS,jiang-etal-2025-tabdsr,jiang2026accuratetablequestionanswering}. 
Some also incorporate memory modules to store and reuse past reasoning experiences \cite{Bai2025MAPLEMA, gu-etal-2025-toward,cheng2026tableminduncertaintyawareprogrammaticagent}. 

As for the tooling part, models generate and execute Python \cite{cao-etal-2023-api, zhang-etal-2024-e5,Yu2025TableRAGAR,pyo2026reasoningcommentedcodetable} or SQL \cite{Abhyankar2024HSTARLH, Nahid2024TabSQLifyER,Khoja2025WeaverIS,thanga2025evidenceguidedschemanormalizationtemporal,sui-etal-2025-chain} code to obtain reasoning results.
To improve code generation, error feedback can be provided to the LLM as a revision signal \cite{cheng-etal-2024-call,Gude2025LySAS, Site2025ITUNLPAS}. To alleviate the reliance on keyword matching inherent in code-based approaches, \citet{nguyen2025improving} propose a framework that leverages LLMs with entity-oriented search. Other commonly used tools include calculator \citep{10.1145/3677052.3698685} and Wikipedia search \citep{Zhou2024EfficientMC}.

In terms of prompting strategies, agentic-flow systems often adopt ReAct-style prompting \cite{Zhou2024EfficientMC, Yu2025TableRAGAR, Bai2025MAPLEMA,WANG2026104723}. 
\citet{Zhang2025RoTET} show that prompting models to iterate over rows can reduce hallucination, while \citet{Dixit2025NoUP} find that no single prompting technique consistently outperforms others in TQA involving temporal reasoning.
\citet{yu2025syntheticdatadrivenprompttuning} propose a self-evolving framework for prompt optimization.

Verification modules have also been introduced to check the correctness of intermediate reasoning \cite{Wang2024AccurateAR, yu-etal-2025-table, luo2026tabtracermontecarlotree,hyeon2026matamultiagentframeworkreliable,tang-etal-2026-exploring}. For instance, \citet{tang-etal-2026-exploring} train process reward models to assess the correctness of each reasoning step, and adopt Best-of-$N$ selection to identify the final answer from multiple sampled reasoning chains. Other similar methods involving test-time scaling include self-consistency \citep{liu-etal-2024-rethinking} and uncertainty-aware sampling \citep{cheng2026tableminduncertaintyawareprogrammaticagent}.
\paragraph{Discussion.} Tuning-free methods require no training data, but incur longer inference times and higher token costs \cite{Zhou2025PPTAP}. In contrast, tuning-based approaches are more efficient at inference but are prone to out-of-domain performance degradation \cite{deng-mihalcea-2025-rethinking, deng-etal-2026-really}. Both methods demonstrate state-of-the-art performance \cite{Yang2025TableR1IS, abhyankar-etal-2025-h, Cao2025TableMasterAR}, but involve trade-offs. A promising intermediate strategy might be to fine-tune models for general reasoning capabilities while delegating specific table operations, such as retrieval, to tools \cite{Wu2024ProTrixBM,10.1145/3677052.3698685}. 

\subsection{Large Inputs}
\label{sec:large-inputs}
The main challenge with large inputs is efficiently identifying relevant information, as processing full tables is often infeasible or ineffective. We review methods for handling large and multiple tables.
\paragraph{Large Tables.}
 A common approach is to fine-tune retrievers to identify the most relevant cells for a given question \cite{lin-etal-2023-inner, Lee2024PieceOT, Lee2024LearningTR, kojima-2024-sub, Patnaik2024CABINETCR}. Another strategy leverages LLMs to select pertinent table content or relevant headers \cite{10.1145/3539618.3591708, jiang-etal-2023-structgpt, Jiang2024SeekAS, sui-etal-2024-tap4llm, sun2026fgtrfinegrainedmultitableretrieval}. Some methods define atomic operations for table manipulation \cite{10942246, Wang2024ChainofTableET}, while others embed both queries and tables for semantic matching \cite{sui-etal-2024-tap4llm, Yu2025TableRAGAR, shirafuji2025hybridsearchcomplextable}. A further line of work employs code generation to execute table-filtering operations \cite{gemmell-dalton-2023-toolwriter, Zhou2025RITTAR, 11060118}.

\paragraph{Multiple Tables.}
LLMs can assist retrieval by enhancing table semantics. For instance, \citet{liang-etal-2025-improving-table} augment table snippets with LLM-generated questions to produce richer table representations.
LLMs can also directly facilitate retrieval. \citet{Kong2024OpenTabAL} generate SQL queries to identify relevant tables. 
Rather than treating each table as an independent document, \citet{Zou2025GTRGF} represent the table corpus as a hypergraph and select the most relevant subgraph using a multi-stage coarse-to-fine process.
Similarly, \citet{luo2026decompositiondrivenmultitableretrievalreasoning} capture table relationships using graphs. They design a question decomposer and a coverage-aware retriever to identify relevant tables.
Many works adopt dense passage retrieval \cite{karpukhin-etal-2020-dense} or language model embeddings
to encode tables, passages, and questions \cite{Guan2024MFORTQAMF, Bardhan2024TTQARSAB}. 
\paragraph{Discussion.} Directly using LLMs to retrieve relevant cells or tables can lead to information loss \cite{Zhou2025RITTAR}. Fine-tuned sub‑table retrievers have shown effectiveness, but their generalizability to diverse table formats remains limited. For both scenarios, retrieval‑augmented generation (RAG) offers a viable alternative: tables or cells are embedded into a vector database, and questions are issued as queries \cite{Chen2024TableRAGMT}.

\subsection{Data Heterogeneity}
\label{sec:data-heterogeneity}
To handle different modalities, existing methods typically follow two directions: (1) employing specialized retrievers and reasoners \cite{li-etal-2022-mmcoqa, 10.1145/3664647.3681053, hu-etal-2024-ket, Liu2023MMHQAICLMI}, or (2) designing unified representations \cite{10.1145/3616855.3635777, Chen2025PandoraAC, 10.1609/aaai.v39i24.34787,park-etal-2025-helios}. In the first category, \citet{hu-etal-2024-ket} use a multi‑stage knowledge‑graph retriever, \citet{10.1145/3664647.3681053} employ multi‑agent retrieval from charts and tables, and \citet{Liu2023MMHQAICLMI} generate image captions to capture salient visual content; atomic retrieval functions can also target both passages and tables \cite{shi-etal-2024-exploring, Zhou2024EfficientMC}. In the second category, unified structures integrate heterogeneous sources: \citet{Chen2025PandoraAC} use DataFrames to jointly represent tabular and textual data, \citet{10.1609/aaai.v39i24.34787} propose Condition Graphs combining tables and knowledge graphs, and \citet{Agarwal2025HybridGF} build hybrid graphs from linked entities; tabular content may also be summarized into text for downstream tasks \cite{Bardhan2024TTQARSAB}. For reasoning over both tables and text, LLMs can align references across modalities \cite{Luo2023HRoTHP, zhang-etal-2025-scitat, sharafath2026n2ngqanoisetonarrativegraphbasedtabletext} or be fine‑tuned for domain‑specific reasoning, as in \citet{10.1145/3677052.3698685}’s financial QA system, which processes both modalities to produce multi‑step reasoning chains combining evidence extraction, logical or equation formulation, and execution.
In summary, which method to choose depends on the modalities involved. Constructing graphs is straightforward for tables and text, whereas employing separate retrievers is preferable when modalities are harder to unify, e.g., if information from charts and tables needs to be combined.


\subsection{Knowledge Integration}
\label{sec:knowledge-integration}
External knowledge is often required to answer TQA problems. For example, a question in DataBench \cite{oses-grijalba-etal-2024-question} asks: \textit{``What is the total number of rebounds recorded in the dataset where the ball didn't change possession?”} Answering this question requires knowing that \textit{OREB} in the table header denotes \textit{offensive rebounds}, a case where ball possession does not change.
To handle such cases, prior work has integrated external resources such as Wikipedia into the reasoning process \cite{Zhou2024EfficientMC, sui-etal-2024-tap4llm}. For instance, \citet{Zhou2024EfficientMC} design an atomic function \textit{Search (arg)}, which returns the first few lines from the Wikipedia page of a specified argument. The retrieved content is then stored in the system’s memory for subsequent reference.
An alternative strategy is to elicit factual knowledge directly from LLMs \cite{Cheng2022BindingLM, Liu2024AugmentBY}. However, this approach is susceptible to hallucinations, as LLMs may generate factually incorrect information.

\section{Evaluation}
\label{sec:evaluation}
In this section, we discuss evaluation in terms of task performance, system  robustness, and model-generated reasoning  as explanations. 

\subsection{Task Performance}
Current TQA evaluation primarily focuses on performance, measured by automatic metrics such as Exact Match (EM) and ROUGE. EM suits short, span-based answers, whereas ROUGE, BLEU, or F1 scores are better for free-form responses. While efficient, these metrics often miss subtle mismatches between predicted and gold answers \cite{Wolff2025HowWD}, e.g., EM may wrongly mark answers as incorrect due to 
formatting differences (Jan 1 vs. 01-01).
To address such issues, outputs are normalized to a canonical form before applying EM \cite{Khoja2025WeaverIS}. This ``relaxed EM” improves robustness but can cause inconsistencies, as systems may adopt different normalization rules, leading to misleading cross-system comparisons \cite{HormazbalLagos2025ExpliCITQAEC}.

Beyond traditional metrics, some studies employ LLMs as judges \cite{Wu2024ProTrixBM, Zhou2024EfficientMC, Jiang2025MultimodalTR, zhang-etal-2025-tablellm} or use human evaluation \cite{zhao-etal-2024-tapera, Ye2024DataFrameQA, Khoja2025WeaverIS}. \citet{Dixit2025NoUP} propose the Hybrid Correctness Score, combining $F_1$ with LLM judgments.
\citet{Wolff2025HowWD} show that, when calibrated with human annotations, LLM-as-a-judge can offer a reliable evaluation signal for tasks requiring reasoning over tables.

\subsection{Robustness Evaluation} 
Robustness to structural or content variations in tables and questions is a key property of TQA systems \citep{yang-etal-2022-tableformer,bhandari2025exploring}. \citet{zhao-etal-2023-robut} present a benchmark for adversarial attacks on table structure/content and question perturbations, showing that state-of-the-art models still struggle. \citet{zhou-etal-2024-freb} define three robustness dimensions: (1) resilience to table structure changes, (2) resistance to shortcut exploitation, and (3) robustness in numerical reasoning. Their benchmark indicates that pipeline models handle value and positional changes best, whereas LLM-based models are more vulnerable to table shuffling, a trend also observed in other works \cite{AshuryTahan2025TheMT, liu-etal-2024-rethinking, yang-etal-2022-tableformer}. \citet{Wolff2025HowWD} further evaluate LLM robustness on real-world tables with missing or duplicated values, underscoring the need for robustness-oriented evaluation.

\subsection{Evaluating Explanations and Reasoning}
An underexplored aspect of TQA evaluation is assessing explanations and reasoning processes. Model-generated chains of thought \cite{Wei2022ChainOT} are often treated as natural language explanations \cite{zhao-etal-2024-tapera, zhou-etal-2025-g, lu-etal-2025-tart}, either elicited directly \cite{zhao-etal-2024-tapera, zhou-etal-2025-g} or derived from executable program outputs \cite{lu-etal-2025-tart}. \citet{Nguyen2024InterpretableLT} represent explanations as chains of attribution maps, showing intermediate relevant tables alongside reasoning steps, and propose three evaluation tasks: (1) preference ranking, where judges rank explanation quality; (2) forward simulation, where judges answer using only the explanation; and (3) verification, where judges assess prediction correctness based on the explanation. Both human annotators and LLMs serve as judges. \citet{Zhou2025PPTAP} take a different approach, estimating the probability of reaching the correct answer from a reasoning step to quantify each step’s contribution to the final outcome.
\paragraph{Discussion.}
Popular datasets like WTQ \cite{pasupat-liang-2015-compositional} and TabFact \cite{Chen2019TabFactAL} may suffer from data contamination \cite{zhou-etal-2025-texts}, leading to overly optimistic performance estimates. \citet{wang2026gleangroundedlightweightevaluation} design a light-weight evaluation protocol taking the issues into consideration. Beyond task performance, dimensions such as robustness and reasoning correctness should be systematically assessed to ensure the development of trustworthy TQA systems.

\section{Discussion and Future Directions}
\label{sec:discussions}
We discuss emerging topics for future exploration, including table representation, multilinguality, reinforcement learning, multi-modal modeling,  interpretability and human-centric setups.

\paragraph{Table Representation.}
Tables appear in various formats, including structured databases, plain text, images, and graphs. When tables are stored in databases, they can be directly queried using SQL. \citet{zhang2026same} find that modeling tables as databases achieves better results than using LLMs directly when tables feature clear schemas. However, tables in textual or image form are often noisier due to inconsistent formatting \cite{zhou-etal-2024-freb}, mixed data types \cite{Nahid2024NormTabIS}, missing values \cite{Wolff2025HowWD}, and implicit or incomplete schemas \cite{zheng-etal-2023-im}, posing challenges for building models that generalize to real-world noisy tables.

A growing line of work investigates leveraging both textual and visual representations of tables \cite{deng-etal-2024-tables, zhou-etal-2025-texts, borisova-etal-2025-table, Liu2025HIPPOET, xing2026tabledart}, capitalizing on the fact that one modality can often be converted to the other (e.g., image-to-text via OCR, or text-to-image via HTML rendering). However, most existing approaches adopt ensemble strategies that select the optimal representation based on specific problem features such as table size \cite{deng-etal-2024-tables, zhou-etal-2025-texts}, which lack flexibility when new features emerge or when interactions among multiple features need to be considered.

To overcome this limitation, some work instead lets the model learn which modality to use \citep{Liu2025HIPPOET, xing2026tabledart}. For instance, \citet{xing2026tabledart} train a gating network to select among text-based, vision-based, or fused representations.
An alternative approach is to process textual and visual inputs through dedicated encoders and integrate them within the model, a strategy widely adopted in vision-language tasks \cite{Zhao2024TabPediaTC, 11093154}. Separate encoders can capture complementary signals, such as layout structure from images and semantic content from text, potentially leading to more robust and generalizable representations.


\paragraph{Multilinguality and Low-Resource Settings.}Tables in real-world applications can be text-heavy and may contain content in one or more languages, as in user or product information tables. Nevertheless, most existing TQA datasets and studies focus on English \cite{pasupat-liang-2015-compositional, zhang-etal-2023-crt, zhang-etal-2024-tablellama} or other high-resource languages with large speaker populations, such as Chinese \cite{zheng-etal-2023-im, liu-etal-2023-tab, 10.1145/3664647.3681053}, leaving low-resource and multilingual scenarios underexplored.
These settings introduce additional challenges, including right-to-left text processing for languages such as Persian and Hebrew, as well as out-of-distribution lexical patterns for LLMs predominantly trained on high-resource languages. Although modern LLMs support multilingual processing, directly applying them to low-resource table reasoning tasks typically results in degraded performance. For instance, \citet{Shu2025M3TQAMM} construct a multilingual table reasoning dataset and report that LLMs achieve their highest performance on Indo-European languages and their lowest on Niger-Congo languages, likely reflecting disparities in pre-training data distribution. Their study also finds that multilingual fine-tuning does not consistently improve table question answering performance. Translation-based approaches, which convert tables from the target language into English prior to reasoning, are similarly limited, as their effectiveness depends heavily on translation quality \citep{Zhang2025MULTITATBM}.

Addressing table understanding and reasoning in multilingual and low-resource contexts is both practically important and ethically motivated: such tasks arise frequently in real-world applications, and equitable digital access requires robust performance across languages. Recent efforts have begun to introduce datasets tailored to these scenarios \citep{minhas-etal-2022-xinfotabs, Zhang2025MULTITATBM, Shu2025M3TQAMM}, but the field remains nascent. We argue that effective table reasoning systems for multilingual and low-resource settings will require more than translation pipelines and off-the-shelf LLMs, calling for dedicated methods that explicitly address the unique linguistic and resource constraints of these environments.

\paragraph{Reinforcement Learning in TQA.} 
\label{rl}

Reinforcement learning with verifiable rewards (RLVR) 
has gained increasing attention due to its success in developing reasoning-oriented models such as DeepSeek-R1 \cite{DeepSeekAI2025DeepSeekR1IR}. Recent studies have also explored RLVR in TQA \cite{Jin2025Tabler1SA, Yang2025TableR1IS, Jiang2025MultimodalTR, Lei2025ReasoningTableER, Cao2025FortuneFR, Stoisser2025SparksOT, Liu2025HIPPOET}. Common rewards include answer correctness \cite{Yang2025TableR1IS, Jiang2025MultimodalTR, Stoisser2025SparksOT, Liu2025HIPPOET}, program executability \cite{Jin2025Tabler1SA, Cao2025FortuneFR}, output formatting \cite{Yang2025TableR1IS, Jiang2025MultimodalTR}, positional alignment \cite{Lei2025ReasoningTableER}, and length constraints \cite{Jin2025Tabler1SA}. \citet{10.1145/3773966.3777932} incorporate tool use into the reward design and propose RAPO, a rank-aware policy optimization algorithm that modifies GRPO by removing KL divergence and employing token-level policy gradients to improve training stability.

Beyond final outcome rewards, several studies explore richer supervision signals. \citet{Zhou2025PPTAP} propose a process supervision framework and demonstrate that models trained with intermediate rewards outperform those trained on final rewards alone.\citet{yang2025tablegptr1advancingtabularreasoning} similarly use process rewards to update policy models. \citet{zhao2026strongermas} employ Monte Carlo Tree Search to generate pseudo-gold trajectories for optimizing agents with reinforcement learning. For table image understanding, \citet{kang-etal-2025-grpo} apply continuous Tree-Edit-Distance Similarity rewards to train models for recognizing table structures and contents.

RLVR has been shown to improve LLMs' reasoning capabilities over tabular data. Compared to supervised fine-tuning (SFT), RLVR-trained models exhibit better generalizability \cite{Yang2025TableR1IS, Cao2025FortuneFR} and greater robustness to row and column perturbations \cite{Lei2025ReasoningTableER}. Nevertheless, initializing models with SFT prior to RLVR training remains crucial for achieving strong performance \cite{Cao2025FortuneFR}.

\paragraph{Diverse and Multi-Modal Data Modeling.}
Many existing TQA studies focus on table-only settings with relatively simple queries \cite{10.1145/3539618.3591708, 10.5555/3618408.3619494, Nahid2024TabSQLifyER, Wang2024ChainofTableET, liu-etal-2024-rethinking, Zhou2025RITTAR}, a setup well-suited for evaluating LLMs' ability to understand table structures. However, it falls short in capturing more complex scenarios that involve multiple modalities and open-domain settings. An increasing body of work has recognized these limitations and proposed more challenging benchmarks \cite{He2023Text2AnalysisAB, Qiu2024TQABenchEL, wu2025mmqa, oses-grijalba-etal-2024-question, Wu2024TableBenchAC, wu-etal-2025-realhitbench, Zhu2025TableEvalAR}.
Beyond information retrieval and aggregation, real-world table interactions are considerably more diverse. Queries can be ambiguous \citep{grijalba-etal-2026-problem}, and users also engage in tasks such as interpreting markup tables \citep{wang2026highlightbenchbenchmarkingmarkupdriventable} and performing table manipulations \citep{xing2025mmtu}. We argue that future research should extend modeling to these real-world scenarios.

\paragraph{Interpretability and Faithfulness.}
Instead of directly producing a final answer \cite{herzig-etal-2020-tapas, Liu2021TAPEXTP, jiang-etal-2022-omnitab, zhang-etal-2025-tablellm}, an increasing number of TQA approaches also return a reasoning process \cite{zhao-etal-2024-tapera, chegini-etal-2025-repanda, Nguyen2024InterpretableLT}. Such reasoning not only improves system performance but also provides human-understandable justifications for how an answer is derived. However, despite appearing plausible, these explanations may not faithfully represent the model's actual decision-making process \cite{Turpin2023LanguageMD, Chen2025ReasoningMD}.
Building trustworthy TQA systems is particularly important in high-stakes domains such as medicine \cite{bardhan-etal-2022-drugehrqa}. Achieving this requires that output reasoning accurately reflect a model's table understanding capabilities, e.g., faithfully responding with ``I don't know'' when a table is beyond the model's ability to interpret. We argue that much of the current work on interpretability in TQA focuses on generating post-hoc justifications for answers, rather than explanations that transparently reveal the underlying reasoning process.

A complementary direction involves probing the internal mechanisms by which LLMs process tabular data. \citet{zhang2026languagemodelsunderstandtables} find that LLMs go through three stages when retrieving table cells: semantic binding (aligning the query with table headers), coordinate localization (navigating structural indices), and information extraction (retrieving target values). During coordinate localization in particular, models locate the target cell by counting discrete delimiters. \citet{wang2026closerlookllmstable} extend this line of inquiry from cell retrieval to general TQA, analyzing model attention patterns during question answering. They find that early layers broadly encode table structure, middle layers localize relevant cells, and later layers generate the final answer.


\paragraph{Human-Centric and Socially-Aware Setups.}
Current research in TQA has largely focused on improving system performance, often overlooking the role of human interaction. However, the ultimate aim of such systems is to empower humans to more effectively interact with tabular data, analogous to the broader goal of developing NLP systems for human and socially aware uses \cite{hovy-yang-2021-importance,ziems-etal-2024-large,yang-etal-2025-socially}. This highlights the importance of human-centric and socially aware design. 
We identify two complementary research directions: (1) human-centric modeling, which emphasizes representations and benchmarks that capture the unique characteristics of tabular data \cite[e.g.,][]{hu-etal-2024-ket,ahmad2025hctqabenchmarkquestionanswering}; and (2) socially grounded applications, where TQA systems serve downstream tasks for social good, such as analyzing environmental sustainability reports \cite{dimmelmeier-etal-2024-informing,Beck2025}, advancing biomedical research \cite{luo2022bio}, and supporting decision-making in financial domains \cite{strich-etal-2026-t2,malarkkan2026finrulebenchbenchmarkjointreasoning}.

\section{Conclusion}
In this survey, we have reviewed recent advances in TQA  with (M)LLMs, covering representative TQA setups, key challenges, and corresponding solutions, along with a discussion of promising future research directions. 
Looking ahead, we envision a stronger synergy between methods from NLP and related fields, and anticipate that TQA research, both in modeling and evaluation, will increasingly adopt more comprehensive settings. Such settings should account for diverse factors, including human model interaction, system robustness, and real-world applicability.
\section*{Limitations}  
In this study, we present a survey on TQA with LLMs. Related surveys are discussed in Appendix \ref{compare_survey}. Despite our best efforts, certain limitations remain.  
  
\paragraph{References and Methods.}    
We collected papers published before April 2026, which means that works appearing after this time are not included. The majority of papers were retrieved from venues such as ACL, EMNLP, NAACL, NeurIPS, ICLR, and arXiv, using English-language queries. This approach may have led to the omission of works published in other languages. Furthermore, due to space constraints, we are unable to provide exhaustive technical details for all methods covered in this survey.  
  
\paragraph{Survey Scope.}    
This survey focuses exclusively on table question answering. Our objective is to provide a comprehensive and detailed overview of the task’s characteristics, challenges, corresponding modeling methods, as well as emerging topics for future research. This focus excludes other table-related tasks, such as table generation and summarization.

\bibliography{custom}

\appendix

\section{Appendix}
\label{sec:appendix}

\subsection{Comparing with Other Surveys}
\label{compare_survey}
We compare our survey with recent work on table question answering (TQA) in Table \ref{tab:compare_survey}. 
\textit{TQA Setups} indicates whether a given survey proposes a TQA-specific task taxonomy or discusses multiple task setups within TQA. 
\textit{Input Modality} specifies the types of input modalities considered in TQA. 
\textit{Lan} denotes the languages of the benchmarks covered in the survey, and \textit{Eval} indicates whether evaluation methodologies are discussed. \textit{RLVR} and \textit{ITPT} refer to reinforcement learning with verifiable rewards and interpretability, respectively. 
As shown in Table \ref{tab:compare_survey}, our survey differs from prior work in the following ways: (1) We provide a fine-grained discussion of diverse TQA task setups and the modalities involved. (2) We include benchmarks covering languages beyond English. (3) We present an up-to-date review of modeling approaches with (M)LLMs and evaluation paradigms. (4) We discuss recent advances and emerging themes in the era of LLMs.

\subsection{Survey Scope}
\label{survey_detail}
We clarify the scope of this survey and describe the process used to collect the papers reviewed.

\paragraph{Tasks.}
This survey primarily focuses on TQA. We also include table fact verification, as this task can be readily reformulated into TQA. For example, by appending a question such as “Is the statement true or false?” to the statement being validated. Another related task is text-to-SQL, for which we mainly discuss relevant datasets, since they can also serve as benchmarks for TQA. We further consider SQL generation as an approach to solving TQA, but we do not provide a detailed review of methods specifically targeting SQL generation. Tasks that are not explicitly covered in this survey include table prediction, table generation, and table summarization, as these differ from TQA in terms of problem formulation and objectives.

\paragraph{Models.}
Given our focus on TQA in the era of LLMs, we primarily include work that leverages these models. For modeling papers, we restrict our collection to works published after 2022, as earlier studies are comprehensively reviewed in the survey by \citet{jin2022surveytablequestionanswering}.

\begin{figure}
\begin{subfigure}{.52\columnwidth}
  \centering
\includegraphics[width=1\columnwidth]{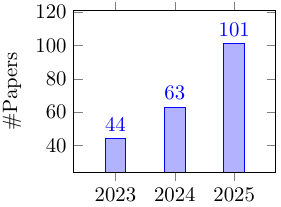}
  \caption{\#collected paper per year.}
  \label{fig:sfig1}
\end{subfigure}%
\begin{subfigure}{.5\columnwidth}
  \centering
  \includegraphics[width=1\columnwidth]{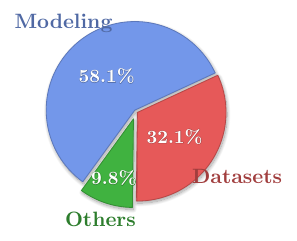}
  \caption{Paper types distribution.}
  \label{fig:sfig2}
\end{subfigure}
\caption{Statistics of the collected paper. We show the number of collected paper by year as well as the distribution of different types of paper.}
\label{fig:survey_statistics}
\end{figure}

\paragraph{Paper Collection.}
We search for papers using the keywords table/tabular reasoning, table/tabular question answering, and table/tabular understanding on both arXiv and the ACL Anthology. We include works published up to April 1st, 2026 and expand the collection by identifying relevant papers cited in the related work sections of the retrieved publications. In total, we compile a corpus of 277 papers. Figure \ref{fig:survey_statistics} presents the distribution of papers by year and theme, illustrating a clear upward trend in research on table question answering.

\begin{table*}[htbp]
    \small
    \centering
     \setlength{\tabcolsep}{1pt}
    \small
\resizebox{1\textwidth}{!}{
    \begin{tabular}{lccccccccl}
         \toprule
         Survey & \makecell{Publication\\ Year} & \makecell{TQA\\ Setups}&\makecell{Input\\Modality}& Lan& \makecell{Modeling \\with LLMs} & Eval &RLVR& ITPT& Summary \\\midrule

         \citet{ijcai2022p761} & 2022 &\textcolor{red}{\xmark} & text& en & \textcolor{red}{\xmark} & \textcolor{red}{\xmark} & \textcolor{red}{\xmark} &   \textcolor{red}{\xmark}& table-pretraining \\
        \citet{jin2022surveytablequestionanswering} & 2022&\textcolor{green}{\cmark}&text&en&\textcolor{red}{\xmark}&\textcolor{red}{\xmark}&\textcolor{red}{\xmark} &  \textcolor{red}{\xmark} & benchmarks \& pre-LLM modeling  \\
        \citet{zhang2024surveytablereasoninglarge}&2024 & \textcolor{red}{\xmark}& text & en &\textcolor{green}{\cmark} & \textcolor{red}{\xmark} &  \textcolor{red}{\xmark} &  \textcolor{red}{\xmark} & LLM-based modeling\\
         \citet{fang2024largelanguagemodelsllmstabular} & 2024 & \textcolor{red}{\xmark} & text & en &\textcolor{green}{\cmark}& p & \textcolor{red}{\xmark} &\textcolor{green}{\cmark}& table prediction, generation and understanding\\

        \citet{Lu2024LargeLM} & 2024 & \textcolor{red}{\xmark} & text, image & en & \textcolor{green}{\cmark} & p,r & \textcolor{red}{\xmark}& \textcolor{red}{\xmark} & tasks discussed via a data lifecycle aspect\\

        \citet{wu2025tabulardataunderstandingllms} &2025  & \textcolor{red}{\xmark} & text, image & en& \textcolor{red}{\xmark}  & p,r & \textcolor{red}{\xmark}  & \textcolor{red}{\xmark} & table representations and tasks  \\
        \citet{tian2025realworldtableagentscapabilities} & 2025 & \textcolor{red}{\xmark} & text, image & en & \textcolor{green}{\cmark} & p,r & \textcolor{red}{\xmark}& \textcolor{red}{\xmark} & LLM agents \\

        \citet{Ren2025DeepLW} & 2025 & \textcolor{red}{\xmark} & text & en & \textcolor{green}{\cmark} &\textcolor{red}{\xmark} & \textcolor{red}{\xmark} & \textcolor{red}{\xmark} & general table modeling with deep learning\\ 

        \citet{XU2026114459} & 2026 &\textcolor{red}{\xmark} & text, image& en& \textcolor{green}{\cmark}&p &\textcolor{green}{\cmark}&\textcolor{green}{\cmark}& TQA from semantic parsing to PLM and to LLM\\
        
        Ours &2026 &   \textcolor{green}{\cmark} &text, image, KB & various  & \textcolor{green}{\cmark} & p,r,e & \textcolor{green}{\cmark} &  \textcolor{green}{\cmark} & \makecell[l]{fine-grained TQA task taxonomy, \\up-to-date modeling and discussion} \\
    \bottomrule
         
    \end{tabular}
}
    \caption{Comparing this work with recent surveys pertinent to table question answering from various perspectives. \textit{TQA Setups}: if a fine-grained TQA task taxonomy is given. \textit{Lan}: language of sourced benchmarks. \textit{Eval}: Evaluation discussions. p,r,e stands for performance, robustness and explanation, respectively. \textit{RLVR}: reinforcement learning with verifiable reward. \textit{ITPT}: interpretability. }
    \label{tab:compare_survey}
\end{table*}


\begin{figure}
    \centering
    \includegraphics[width=0.8\columnwidth]{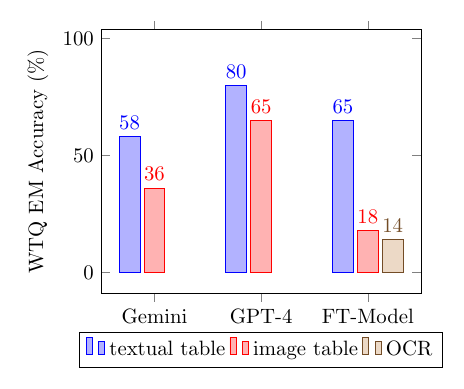}
    \caption{Performance of (M)LLMs in textual and image-based table understanding. FT-Model denotes fine-tuned models, specifically TableLlaVA-7B \cite{zheng-etal-2024-multimodal} and TableLlaMA-7B \cite{zhang-etal-2024-tablellama}. OCR refers to configurations in which image tables are first converted to text via optical character recognition (OCR) and then processed using TableLlaMA-7B.}
    \label{fig:compare_text_image}
\end{figure}
\subsection{Method Comparison}
\label{method_compare}
Figure \ref{fig:compare_text_image} compares the performance of (multimodal) large language models ((M)LLMs)) in textual and image-based table understanding. The results are drawn from \citet{deng-etal-2024-tables} and \citet{zheng-etal-2024-multimodal}.

\subsection{TQA Datasets}
Table \ref{tab:mapping_task_dataset_full} shows existing TQA datasets, categorized by features introduced in Section \ref{sec:task_dimensions}. Licenses are subject to each individual dataset. Please refer to the original datasets for more information.
\label{tqa_datasets}

\begin{table*}[htbp]
    \footnotesize
    \centering
     \setlength{\tabcolsep}{2pt}
    \small
    \caption{Existing TQA Datasets. \textit{MC} denotes multiple-choice. \textit{ret} and \textit{rea} refer to retrieval and reasoning, respectively. \textit{LAN} indicates the language of the dataset. \textit{Q} and \textit{T} represent question and table, respectively. The value \textit{both} in the table suggests that the dataset contains both single and multiple tables, or includes both flat and hierarchical tables.\textit{DB} stands for database.}
\label{tab:mapping_task_dataset_full}
\resizebox{1\textwidth}{!}{
    \begin{tabular}{ccccccccccccccc}
         \toprule
        \multirow{2}{*}{Datasets}&\multirow{2}{*}{\makecell{Source/\\Domain}}&\multirow{2}{*}{\makecell{Open\\domain}}&\multirow{2}{*}{\makecell{Answer\\format}} &\multirow{2}{*}{Reasoning}
        &\multirow{2}{*}{LAN}&\multicolumn{2}{c}{Size}&\multicolumn{3}{c}{table features}&\multicolumn{4}{c}{additional inputs}  \\ 
  \cmidrule(l){7-8} \cmidrule(l){9-11} \cmidrule(l){12-15}
&&&&&&\#Q&\#T&Single&Flat&Format&text&image&chart&KB\\
          \midrule
         \makecell{WTQ \\\cite{pasupat-liang-2015-compositional}}&Wikipedia&\xmark& spans&ret, rea& EN&22k&2.1k&\cmark&\cmark&text&\xmark&\xmark&\xmark&\xmark\\

      \makecell{SQA \\\cite{iyyer-etal-2017-search}}&WTQ&\xmark& spans&ret, rea &EN&17k&&\cmark&\cmark&text&\xmark&\xmark&\xmark&\xmark\\
      
      \makecell{TabMCQ \\\cite{jauhar-etal-2016-tables}}&science exam&\xmark& MC& &EN&9k&68&\cmark&\cmark&text&\xmark&\xmark&\xmark&\xmark\\

      \makecell{TabFact \\\cite{Chen2019TabFactAL}}&Wikipedia&\xmark& MC&ret, rea &EN&118k&16k&\cmark&\cmark&text&\xmark&\xmark&\xmark&\xmark\\

      \makecell{WikiSQL \\\cite{Zhong2017Seq2SQLGS}}&Wikipedia&\xmark& SQL&ret, rea &EN&80k&26k&\cmark&\cmark&text&\xmark&\xmark&\xmark&\xmark\\

      \makecell{Spider \\\cite{yu-etal-2018-spider}}&\makecell{WikiSQL,\\Internet}&\xmark& SQL&ret, rea &EN&10k&200&both&\cmark&DB&\xmark&\xmark&\xmark&\xmark\\

      \makecell{INFOTABS \\\cite{gupta-etal-2020-infotabs}}&Wikipedia&\xmark& MC&ret, rea &EN&23k&2.5k&\cmark&\cmark&text&\xmark&\xmark&\xmark&\xmark\\

      \makecell{KaggleDBQA  \\\cite{lee-etal-2021-kaggledbqa}}&Kaggle&\xmark&SQL&ret,rea&EN&272&8&both&\cmark&DB&\xmark&\xmark&\xmark&\xmark\\

      \makecell{HiTab \\\cite{cheng-etal-2022-hitab}}&\makecell{statistical \\report}&\xmark& spans&ret, rea &EN&10k&3.6k&\cmark&\xmark&text&\xmark&\xmark&\xmark&\xmark\\

      \makecell{AIT-QA \\\cite{katsis-etal-2022-ait}}&airline&\xmark& spans&ret &EN&515&116&\cmark&both&text&\xmark&\xmark&\xmark&\xmark\\

      \makecell{FeTaQA \\\cite{nan-etal-2022-fetaqa}}&ToTTo &\xmark& free-form&ret, rea &EN&10k&10k&\cmark&\cmark&text&\xmark&\xmark&\xmark&\xmark\\

      \makecell{TabMWP \\\cite{Lu2022DynamicPL}}&websites &\xmark& MC, spans&numerical &EN&38k&37k&\cmark&\cmark&\makecell{text,\\image}&\xmark&\xmark&\xmark&\xmark\\

      \makecell{XINFOTABS \\\cite{minhas-etal-2022-xinfotabs}}&INFOTABS &\xmark& MC&ret, rea &MLT&23k&2.5k&\cmark&\cmark&text&\xmark&\xmark&\xmark&\xmark\\

      \makecell{EI-INFOTABS \\\cite{agarwal-etal-2022-bilingual}}&INFOTABS &\xmark& MC&ret, rea &HI&23k&2.5k&\cmark&\cmark&text&\xmark&\xmark&\xmark&\xmark\\

      \makecell{KorWikiTQ \\\cite{jun-etal-2022-korean}}&Wikipedia &\xmark& spans&ret, rea &KO&70k&&\cmark&\cmark&text&\xmark&\xmark&\xmark&\xmark\\

      \makecell{TempTabTQA \\\cite{gupta-etal-2023-temptabqa}}&Wikipedia &\xmark& spans&temporal  &EN&11k&1.2k&\cmark&\cmark&text&\xmark&\xmark&\xmark&\xmark\\

      \makecell{RobuT \\\cite{zhao-etal-2023-robut}}&\makecell{WTQ, SQA, \\ WikiSQL} &\xmark& spans&robustness&EN&143k&&\cmark&\cmark&text&\xmark&\xmark&\xmark&\xmark\\

      \makecell{IM-TQA \\\cite{zheng-etal-2023-im}}&reports &\xmark& spans&ret,rea&ZH&5k&1.2k&\cmark&\xmark&text&\xmark&\xmark&\xmark&\xmark\\

      \makecell{Text2Analysis \\\cite{He2023Text2AnalysisAB}}& &\xmark& code&ret,rea&EN&2.2k&347&\cmark&\cmark&text&\xmark&\xmark&\xmark&\xmark\\

      \makecell{SciTab \\\cite{lu-etal-2023-scitab}}&SciGen  &\xmark& MC&ret,rea&EN&1.2k&&\cmark&\cmark&text&\xmark&\xmark&\xmark&\xmark\\

      \makecell{CRT \\\cite{zhang-etal-2023-crt}}&TabFact  &\xmark& spans&ret,rea&EN&728&423&\cmark&\cmark&text&\xmark&\xmark&\xmark&\xmark\\

      \makecell{Tab-CQA \\\cite{liu-etal-2023-tab}}&reports  &\xmark& spans&ret&ZH&10k&7k&\cmark&&text&\xmark&\xmark&\xmark&\xmark\\

      \makecell{BIRD \\\cite{10.5555/3666122.3667957}}&internet&\xmark&SQL&ret,rea&EN&12.7k&95&both&\cmark&DB&\xmark&\xmark&\xmark&\xmark\\

      \makecell{LF-TQA \\\cite{wang-etal-2024-revisiting}}&\makecell{FeTAQA, \\ QTSUMM}  &\xmark& free-form&ret,rea&EN&2.9k&&\cmark&\cmark&text&\xmark&\xmark&\xmark&\xmark\\

      \makecell{Indic-TQA \\\cite{pal-etal-2024-table}}&Wikipedia  &\xmark& spans&ret,rea&BN,Hi&2m&21k&\cmark&\cmark&text&\xmark&\xmark&\xmark&\xmark\\

      \makecell{TableBench \\\cite{Wu2024TableBenchAC}}&\makecell{existing \\datasets}  &\xmark& \makecell{spans, \\free-form}&ret,rea&EN&20k&3.6k&\cmark&\cmark&text&\xmark&\xmark&\xmark&\xmark\\

      \makecell{Databench \\\cite{oses-grijalba-etal-2024-question}}&internet&\xmark&spans&ret,rea&EN&1.8k&65&\cmark&\cmark&DB&\xmark&\xmark&\xmark&\xmark\\

      \makecell{DACO \\\cite{Wu2024DACOTA}}&\makecell{Spider,\\ Kaggle}&\xmark&free-form&ret,rea&EN&1.9k&440&\xmark&\cmark&DB&\xmark&\xmark&\xmark&\xmark\\

      \makecell{TabIS \\\cite{pang-etal-2024-uncovering}}&\makecell{ToTTo\\ HiTab}&\xmark&MC&ret,rea&EN&61k&&\cmark&\cmark&DB&\xmark&\xmark&\xmark&\xmark\\

      \makecell{FREB-TQA \\\cite{zhou-etal-2024-freb}}&\makecell{existing\\datasets}&\xmark&spans&robustness&EN&8.5k&&\cmark&\cmark&DB&\xmark&\xmark&\xmark&\xmark\\

      \makecell{RealTabBench \\\cite{Su2024TableGPT2AL}}&\makecell{existing\\datasets}&\xmark&free-form&robustness&EN,ZH&6k&360&\cmark&both&csv&\xmark&\xmark&\xmark&\xmark\\

    \bottomrule
         
    \end{tabular}
}
\end{table*}

\begin{table*}[htbp]
    \footnotesize
    \centering
     \setlength{\tabcolsep}{2pt}
    \small
\resizebox{1\textwidth}{!}{
    \begin{tabular}{ccccccccccccccc}
         \toprule
        \multirow{2}{*}{Datasets}&\multirow{2}{*}{\makecell{Source/\\Domain}}&\multirow{2}{*}{\makecell{Open\\domain}}&\multirow{2}{*}{\makecell{Answer\\format}} &\multirow{2}{*}{Reasoning}
        &\multirow{2}{*}{LAN}&\multicolumn{2}{c}{Size}&\multicolumn{3}{c}{table features}&\multicolumn{4}{c}{additional inputs}  \\ 
  \cmidrule(l){7-8} \cmidrule(l){9-11} \cmidrule(l){12-15}
&&&&&&\#Q&\#T&Single&Flat&Format&text&image&chart&KB\\
          \midrule

      \makecell{TabularMath \\\cite{tian2026tabularmathunderstandingmathreasoning}}&GSM8K&\xmark&number&ret,rea&EN&6.7k&3.3k&\cmark&\cmark&\makecell{text\\image}&\xmark&\xmark&\xmark&\xmark\\

      \makecell{NIAT \\\cite{Wang2025NeedleInATableEL}}&\makecell{WTQ, HiTab\\ AIT-QA}&\xmark&spans&ret&EN&287k&750&\cmark&both&text&\xmark&\xmark&\xmark&\xmark\\

   \makecell{HCT-QA \\\cite{ahmad2025hctqabenchmarkquestionanswering}}&internet&\xmark&spans&ret,rea&EN,AR&77k&6.7k&\cmark&\xmark&\makecell{text\\image}&\xmark&\xmark&\xmark&\xmark\\

      \makecell{TableEval \\\cite{Zhu2025TableEvalAR}}&reports&\xmark&spans&ret,rea&EN,ZH&2.3k&617&\cmark&both&csv&\xmark&\xmark&\xmark&\xmark\\
      
      \makecell{SciAtomicBench \\\cite{Zhang2025AtomicRF}}&\makecell{PubTables\\MatSciTable}&\xmark&MC&ret,rea&EN&2.5k&&\cmark&both&text&\xmark&\xmark&\xmark&\xmark\\

      \makecell{AraTable \\\cite{Alshaikh2025AraTableBL}}&Internet&\xmark&spans&ret,rea&AR&615&41&\cmark&\cmark&text&\xmark&\xmark&\xmark&\xmark\\

      \makecell{RealHiTBench \\\cite{wu-etal-2025-realhitbench}}&\makecell{online\\platforms}&\xmark&free-form&ret,rea&EN&3.7k&708&both&\xmark&\makecell{text\\image}&\xmark&\xmark&\xmark&\xmark\\

      \makecell{TReB \\\cite{Li2025TReBAC}}&\makecell{existing\\datasets}&\xmark&free-form&ret,rea&EN,ZH&7.8k&&\cmark&both&text&\xmark&\xmark&\xmark&\xmark\\

      \makecell{M3TQA \\\cite{Shu2025M3TQAMM}}&reports&\xmark&spans&ret,rea&MLT&46k&50&\cmark&both&text&\xmark&\xmark&\xmark&\xmark\\

      \makecell{TableDreamer \\\cite{zheng-etal-2025-tabledreamer}}&synthesized&\xmark&free-form&ret,rea&EN&27k&&\cmark&\cmark&text&\xmark&\xmark&\xmark&\xmark\\

      \makecell{RUST-BENCH \\\cite{abhyankar2025rustbenchbenchmarkingllmreasoning}}& \makecell{NSF, \\Sportsett} &\xmark&spans&ret,rea&EN&7.9k&2k&\cmark&\cmark&text&\xmark&\xmark&\xmark&\xmark\\

      \makecell{SciTaRC \\\cite{wang2026scitarcbenchmarkingqascientific}}& arXiv &\xmark&free--form&ret,rea&EN&370k&3.7k&\cmark&both&LaTex&\xmark&\xmark&\xmark&\xmark\\

      \makecell{MMQA \\\cite{wu2025mmqa}}&Spider &\xmark&spans&ret,rea&EN&3.3k&3.3k&\xmark&\cmark&DB&\xmark&\xmark&\xmark&\xmark\\

      \makecell{MultiTableQA \\\cite{Zou2025GTRGF}}&\makecell{existing\\datasets} &\xmark&spans&ret,rea&EN&23k&57k&\xmark&\cmark&text&\cmark&\xmark&\xmark&\xmark\\

      \makecell{TRANSIENTTABLES \\\cite{shankarampeta-etal-2025-transienttables}}&Wikipedia &\xmark&spans&temporal&EN&3.9k&14k&\xmark&\cmark&text&\xmark&\xmark&\xmark&\xmark\\

      \makecell{MiMoTable \\\cite{li-etal-2025-mimotable}}&internet &\xmark&free-form&ret,rea&EN,ZH&1.7k&428&both&both&csv&\xmark&\xmark&\xmark&\xmark\\

      \makecell{TQA-Bench \\\cite{Qiu2024TQABenchEL}}&\makecell{world-Bank\\BIRD} &\xmark&MC&ret,rea&EN&56k&10&\xmark&\cmark&DB&\xmark&\xmark&\xmark&\xmark\\

      \makecell{ReasonTabQA \\\cite{pan2026reasontabqacomprehensivebenchmarktable}}&\makecell{Internet} &\xmark&spans&ret,rea&EN,ZH&5.5k&1.9k&\xmark&\xmark&text&\xmark&\xmark&\xmark&\xmark\\

      \makecell{ComTQA \\\cite{Zhao2024TabPediaTC}}&\makecell{FinTabNet\\PubTab1M D} &\xmark&spans&ret,rea&EN&9k&1.5k&\cmark&both&image&\xmark&\xmark&\xmark&\xmark\\

      \makecell{MMTab \\\cite{zheng-etal-2024-multimodal}}&\makecell{existing\\datasets} &\xmark&spans&ret,rea&EN&49k&23k&\cmark&both&image&\xmark&\xmark&\xmark&\xmark\\

      \makecell{MTabVQA \\\cite{Singh2025MTabVQAEM}}&\makecell{existing\\datasets} &\xmark&spans&ret,rea&EN&3.7k&8.5k&\xmark&\cmark&image&\xmark&\xmark&\xmark&\xmark\\

      \makecell{MMSci \\\cite{Yang2025DoesTS}}&SciGen  &\xmark&spans&ret,rea&EN&15k&52k&\cmark&\cmark&image&\xmark&\xmark&\xmark&\xmark\\

      \makecell{TabComp \\\cite{gautam-etal-2025-tabcomp}}& DocVQA  &\xmark&free-form&ret,rea&EN&30k&10k&\cmark&both&image&\xmark&\xmark&\xmark&\xmark\\

      \makecell{TableVQA-Bench \\\cite{Kim2024TableVQABenchAV}}& \makecell{WTQ, TabFact\\ FinTabNet}  &\xmark&spans&ret,rea&EN&1.5k&894k&\cmark&\cmark&image&\xmark&\xmark&\xmark&\xmark\\

      \makecell{TABLET \\\cite{alonso2026tablet}}& \makecell{existing\\ datasets}  &\xmark&spans&ret,rea&EN&4000k&2000k&\cmark&both&image&\xmark&\xmark&\xmark&\xmark\\

      \makecell{MMFCTUB \\\cite{yakun2026mmfctubmultimodalfinancialcredit}}& synthesized &\xmark&MC&ret,rea&EN&25k&19k&\cmark&\cmark&image&\xmark&\xmark&\xmark&\xmark\\

      \makecell{Visual-TableQA \\\cite{lompo2025visualtableqaopendomainbenchmarkreasoning}}& synthesized &\xmark&free-form&ret,rea&EN&6k&2.5k&\cmark&both&image&\xmark&\xmark&\xmark&\xmark\\

      \makecell{CoReTab \\\cite{nguyen-okatani-2026-coretab}}& MMTab  &\xmark&spans&ret,rea&EN&115k&&\cmark&both&image&\xmark&\xmark&\xmark&\xmark\\

      \makecell{OTT-QA \\\cite{Chen2020OpenQA}}& Wikipedia  &\cmark&spans&ret,rea&EN&45k&&&\cmark&text&\cmark&\xmark&\xmark&\xmark\\

      \makecell{TANQ \\\cite{Akhtar2024TANQAO}}& \makecell{QAMPARI\\Wikipedia}  &\cmark&table&ret,rea&EN&43k&&&\cmark&text&\cmark&\xmark&\xmark&\xmark\\

      \makecell{RETQA \\\cite{Wang2024RETQAAL}}& \makecell{QAMPARI\\Wikipedia}  &\cmark&free-form&ret,rea&ZH&20k&4.9k&&\cmark&DB&\cmark&\xmark&\xmark&\xmark\\

      \makecell{Open-WikiTable \\\cite{kweon-etal-2023-open}}& \makecell{WTQ\\WikiSQL}  &\cmark&spans&ret,rea&EN&67k&24k&&\cmark&text&\xmark&\xmark&\xmark&\xmark\\
      
    \bottomrule
         
    \end{tabular}
}    
\end{table*}

\begin{table*}[ht]
    \footnotesize
    \centering
     \setlength{\tabcolsep}{2pt}
    \small
\resizebox{1\textwidth}{!}{
    \begin{tabular}{ccccccccccccccc}
         \toprule
        \multirow{2}{*}{Datasets}&\multirow{2}{*}{\makecell{Source/\\Domain}}&\multirow{2}{*}{\makecell{Open\\domain}}&\multirow{2}{*}{\makecell{Answer\\format}} &\multirow{2}{*}{Reasoning}
        &\multirow{2}{*}{LAN}&\multicolumn{2}{c}{Size}&\multicolumn{3}{c}{table features}&\multicolumn{4}{c}{additional inputs}  \\ 
  \cmidrule(l){7-8} \cmidrule(l){9-11} \cmidrule(l){12-15}
&&&&&&\#Q&\#T&Single&Flat&Format&text&image&chart&KB\\
          \midrule

 \makecell{CompMix \\\cite{Christmann2023CompMixAB}}& CONVMIX  &\cmark&spans&ret,rea&EN&9.4k&&\cmark&\cmark&text&\cmark&\xmark&\xmark&\cmark\\

      \makecell{$T^2$-RAGBench \\\cite{strich-etal-2026-t2}}& \makecell{existing\\datasets}  &\cmark&spans&ret,rea&EN&32k&&&&text&\cmark&\xmark&\xmark&\xmark\\

      \makecell{MMCoQA\\\cite{li-etal-2022-mmcoqa}}& MMQA   &\cmark&spans&ret,rea&EN&5.7k&10k&&&text&\cmark&\cmark&\xmark&\xmark\\

      \makecell{MMTBENCH\\\cite{Titiya2025MMTBENCHAU}}& Internet   &\xmark&spans&ret,rea&EN&4k&500&\cmark&both&\makecell{csv\\image}
      &\xmark&\cmark&\cmark&\xmark\\

      \makecell{KET-QA\\\cite{hu-etal-2024-ket}}& HybridTQA   &\xmark&spans&ret,rea&EN&9.4k&5.7k&\cmark&\cmark&text
      &\xmark&\xmark&\xmark&\cmark\\

      \makecell{CT2C-QA\\\cite{10.1145/3664647.3681053}}& reports   &\xmark&spans&ret,rea&ZH&9.9k&369&&&html
      &\cmark&\xmark&\cmark&\xmark\\

      \makecell{mmtabqa\\\cite{mathur-etal-2024-knowledge}}& \makecell{existing\\datasets}   &\xmark&spans&ret,rea&EN&69k&259&\cmark&\cmark&text
      &\xmark&\cmark&\xmark&\xmark\\

      \makecell{StructFact\\\cite{huang-etal-2025-structfact}}& \makecell{existing\\datasets}   &\xmark&MC&ret,rea&EN&13k&&\cmark&\cmark&text
      &\cmark&\xmark&\xmark&\cmark\\

      \makecell{WikiMixQA\\\cite{Foroutan2025WikiMixQAAM}}& WTabHTML  &\xmark&MC&ret,rea&EN&1k&&both&\cmark&html
      &\cmark&\cmark&\cmark&\xmark\\

      \makecell{SPIQA\\\cite{Pramanick2024SPIQAAD}}& arXiv   &\xmark&free-form&ret,rea&EN&270k&117k&both&both&image
      &\cmark&\cmark&\cmark&\xmark\\

      \makecell{MMQA\\\cite{Talmor2021MultiModalQACQ}}& Wikipedia    &\xmark&spans&ret,rea&EN&30k&&\cmark&\cmark&text
      &\cmark&\cmark&\xmark&\xmark\\

   \makecell{SciTabQA \\\cite{ghosh-etal-2024-robust}}&SciGen &\xmark&spans&ret,rea&EN,AR&822&198&\cmark&\cmark&text&\cmark&\xmark&\xmark&\xmark\\

   \makecell{MULTITAT \\\cite{Zhang2025MULTITATBM}}&\makecell{existing\\datasets} &\xmark&spans&ret,rea&MLT&250&&\cmark&\cmark&text&\cmark&\xmark&\xmark&\xmark\\

   \makecell{SciTAT \\\cite{zhang-etal-2025-scitat}}&arXiv&\xmark&\makecell{spans\\free-form}&ret,rea&EN&13k&&\cmark&\cmark&text&\cmark&\xmark&\xmark&\xmark\\

   \makecell{PACIFIC \\\cite{deng-etal-2022-pacific}}&TATQA&\xmark&spans&ret,rea&EN&19k&&\cmark&\cmark&text&\cmark&\xmark&\xmark&\xmark\\

   \makecell{TAT-QA \\\cite{zhu-etal-2021-tat}}&reports&\xmark&spans&ret,rea&EN&16k&&\cmark&\cmark&text&\cmark&\xmark&\xmark&\xmark\\

   \makecell{GeoTSQA \\\cite{Li2021TSQATS}}&exams&\xmark&MC&ret,rea&ZH&1k&&\xmark&&text&\cmark&\xmark&\xmark&\xmark\\

   \makecell{HybridQA \\\cite{chen-etal-2020-hybridqa}}&Wikipedia &\xmark&spans&ret&EN&70k&13k&\cmark&\cmark&text&\cmark&\xmark&\xmark&\xmark\\

   \makecell{FinQA \\\cite{chen-etal-2021-finqa}}&reports &\xmark&spans&ret,rea&EN&8k&&\cmark&\cmark&text&\cmark&\xmark&\xmark&\xmark\\

   \makecell{MultiHiertt \\\cite{zhao-etal-2022-multihiertt}}&FinTabNet &\xmark&spans&ret,rea&EN&10k&&\xmark&\xmark&text&\cmark&\xmark&\xmark&\xmark\\

   \makecell{SPARTA \\\cite{park2026sparta}}&ROTOWIRE &\xmark&spans&ret,rea&EN&3.3k&&\xmark&\cmark&DB&\cmark&\xmark&\xmark&\xmark\\
   
    \bottomrule
         
    \end{tabular}
}    
\end{table*}

\end{document}